\documentclass{article}

\usepackage{PRIMEarxiv}
\usepackage{amsmath, amsfonts, amsthm, amssymb,epsfig}
\usepackage[utf8]{inputenc} 
\usepackage[T1]{fontenc}    
\usepackage{hyperref}       
\usepackage{url}            
\usepackage{booktabs}       
\usepackage{amsfonts}       
\usepackage{nicefrac}       
\usepackage{microtype}      
\usepackage{lipsum}
\usepackage{fancyhdr}       
\usepackage{graphicx}       
\graphicspath{{media/}} 
\usepackage{subcaption}
\usepackage{algorithm}
\usepackage{algpseudocode}
\title{Video-based Surgical Skill Assessment using Tree-based Gaussian Process Classifier
}

\author{
  Arefeh Rezaei, Mohammad Javad Ahmadi \\
  Advanced Robotics and Automated Systems  (ARAS), Faculty of Computer and Electrical Engineering\\ 
  K. N. Toosi University of Technology \\
  Tehran, Iran\\
  \texttt{a\_rezai@email.kntu.ac.ir, MJAHMADEE@gmail.com}\\
   \And
  Amir Molaei \\  
  Concordia University\\
  Gina Cody School of Engineering and Computer Science\\ 
  Montreal, Quebec, Canada\\
  \texttt{a\_molaei@encs.concordia.ca}\\
   \AND
  Hamid. D. Taghirad \\
  Advanced Robotics and Automated Systems  (ARAS), Faculty of Electrical Engineering\\ 
  K. N. Toosi University of Technology \\
  Tehran, Iran \\
  \texttt{taghirad@kntu.ac.ir} \\
}
\begin{document}
\maketitle

\begin{abstract}
This paper aims to present a novel pipeline for automated surgical skill assessment using video data and to showcase the effectiveness of the proposed approach in evaluating surgeon proficiency, its potential for targeted training interventions, and quality assurance in surgical departments. The pipeline incorporates a representation flow convolutional neural network and a novel tree-based Gaussian process classifier, which is robust to noise, while being computationally efficient. Additionally, new kernels are introduced to enhance accuracy. The performance of the pipeline is evaluated using the JIGSAWS dataset. Comparative analysis with existing literature reveals significant improvement in accuracy and betterment in computation cost. The proposed pipeline contributes to computational efficiency and accuracy improvement in surgical skill assessment using video data. Results of our study based on comments of our colleague surgeons show that the proposed method has the potential to facilitate skill improvement among surgery fellows and enhance patient safety through targeted training interventions and quality assurance in surgical departments.
\end{abstract}


\section{Introduction}
Surgical skill assessment is the process of evaluating a surgeon's technical ability and competence in performing surgical procedures to ensure patient safety and the quality of care delivered. Skill assessments play a crucial role in certifying aspiring surgeons. Such assessments aid in efficient skill acquisition by providing specific guidance on areas that need improvement~\cite{FormativeandSummativeAssessmentNIU}. In the conventional approach, inexperienced surgery fellows receive instruction and appraisal from seasoned professionals in the same field. To guarantee impartial and consistent evaluations, standardized procedure-specific checklists and global rating scales have been implemented for assessing surgical skills~\cite{ahmed2011observational}. Nonetheless, using these evaluation procedures can be costly and time-intensive. To address this issue, the notion of automating the process of surgical skill evaluation has emerged as a promising solution. Not only would this approach save time and resources, but it would also enable novice surgeons to be trained independently, using a surgical simulator that provides automatic assessment and feedback. As a result, there has been a growing interest in the research community on automatic skill evaluation over the past decade.

In automated surgical skill assessment (ASSA), sensory data such as robot kinematics, tool motion, force, or video data are collected during surgical training. This data is subsequently analyzed to determine the trainee's surgical skill level, which can be expressed either numerically as the average score or categorically as novice, intermediate, or expert~\cite{funke2019video}. The commonly used methods in ASSA are neural network (NN), Hidden Markov Model (HMM), and Support Vector Machine (SVM)~\cite{lam2022machine}. HMM can capture the temporal dependencies of the actions and can be trained using annotated surgical videos or physiological signals. In addition, it can capture the underlying patterns of the data and make probabilistic predictions. Unlike NN-based methods, HMM approaches can learn without numerous data. However, tuning the hyper-parameters of the HMM is not straightforward~\cite{ahmidi2015automated,sun2017smart}. SVM can handle non-linear decision boundaries through the use of kernel functions. However, SVMs is sensitive to the choice of kernel function and parameters. The computational complexity of SVM procedures is light, while it is an arduous process for multi-label and massive  datasets~\cite{ershad2019automatic}. According to the state-of-the-art research~\cite{lam2022machine}, the achievable performance accuracy of the SVM for video classification is yet not satisfactory. In addition, CNNs can learn to estimate the locations of surgeons' tools from surgical videos, such as Temporal Segment Networks (TSN) and Faster Region-based CNN (Faster-RCNN)~\cite{wang2016temporal, ren2015faster}. CNNs have also been trained to send meaningful feedback or give scores to surgeons. Although the deep learning-based approaches can accurately perform assessments~\cite{zhang2020automatic,funke2019video}, however, to learn well, they require very rich datasets with highly intricate computations. 

While most ASSA methods primarily analyze robot kinematic or tool motion data, acquiring such data necessitates access to either a robotic surgical system or specialized tracking systems, which makes it complex and costly in practice. In contrast, video data can be effortlessly obtained in both traditional and robot-assisted minimally invasive surgery. However, video data is high-dimensional and much more intricate than sequences of a few motion variables. The recorded video can be used to obtain kinematic data by tracking surgical tools and assess surgical skill by analyzing this data without the requirement of having motion tracking markers~\cite{bouget2017vision}. Such kinematic data can also be obtained using a convolutional neural network (NN). In~\cite{du2018articulated}, the authors propose a deep neural network for articulated multi-instrument 2-D pose estimation, which is trained on detailed annotations of endoscopic and microscopic data sets. The overall scheme can effectively localize instrument joints and also estimate the articulation model. In another approach, video data can be converted to time series for ASSA. In~\cite{zia2018video}, the authors use entropy-based features approximate entropy, and cross-approximate entropy to find the difference in the predictability of motions. This method is based on the fact that an expert surgeon will have more predictable hand motion while a beginner will exhibit erratic and irregular patterns. 

The two above methods discussed require intermediate steps, which brings in complexity. Furthermore, in~\cite{funke2019video}, both spatial and temporal information are captured using CNN for surgical skill assessment. Thus, new studies have been focused on methods that exclusively rely on using surgical videos. In video-based action recognition and skill assessment, some strategies~\cite{lin2021efficient, kitaguchi2021development} demonstrate that utilization of optical flows of videos with RGB videos simultaneously causes further accuracy in classification. For this purpose,~\cite{funke2019video} has rendered a two-streamed CNN, which contains two parallel networks and gets two kinds of inputs: RGB frame sequences and optical flow feature maps, while satisfactory accuracy is acquired, the computations are very intricate. Performed experiments of such networks in surgical skill assessment demonstrate that accuracy has increased compared to that of the trained model with only RGB frames~\cite{li2019manipulation}. Computing optical flows of videos compel us to consume additional time. To overcome this issue,~\cite{piergiovanni2019representation} introduces a unified structure that uses optical flow features, although employing this structure in action recognition has been performed with better precision than two-streamed CNNs. Since this network is very promising, it has been applied in our surgical skill assessment framework. In another study~\cite{li2021real}, a real-time Gaussian process-based approach has been used for real-time ASSA using kinematic data.

Continuing the work of previous studies, we aimed to automatically assess surgical skills directly from video data, with a new architecture referred to as Representation Flow Noisy INput GP-Tree (RF-NIGP-Tree). To this end, we have developed a computationally efficient pipeline for ASSA with limited video data. The proposed method is a CNN-based method with an inherent representation flow (RF) algorithm introduced in~\cite{piergiovanni2019representation}. This method was originally introduced for action recognition, coupled with NIGP-Tree (noisy input GP-Tree ~\cite{achituve2021gp}, which is a GP based classification algorithm (GPC)). NIGP-Tree is an enhanced GP-Tree method robust to noise, introduced in this study, and the result of the analysis shows that NIGP-Tree improves the accuracy of SSA. We also introduce two variants of RBF kernel for GPC, showing that for both GP-Tree and NIGP-Tree, the accuracy will be improved by using them. Compared to the existing method~\cite{funke2019video}, the proposed method relies on less number of frames of videos (short clips) for training. Additionally, the number of frames for test data are fewer compared to that of the training.

The remainder of the paper is structured as follows. In Section~\ref{background}, an overview of the background and preliminaries are presented. Next, in Section~\ref{proposed method}, we discuss the proposed pipeline for SSA. Section~\ref{Results}, provides the results of the proposed method compared to the existing ones. Finally, the concluding remarks are given in Section~\ref{Conclusion}.

\section{Background and Preliminaries}
\label{background}
\subsection{JIGSAWS}
The JHU-ISI Gesture and Skill Assessment Working Set (JIGSAWS)~\cite{gao2014jhu} is a publicly available surgical dataset created by researchers at Johns Hopkins University and the University of Southern California. The dataset contains recordings of surgical tasks performed by both experienced and novice surgeons using the da Vinci Surgical System. The JIGSAWS dataset includes videos of three surgical tasks: needle passing, suturing, and knot tying. The dataset also includes kinematic and tool usage data, such as the position and orientation of the surgical instruments, the amount of force applied by the surgeon, and the speed of the movements. JIGSAWS also contains annotations for surgical gestures and skill, which have been rendered manually, based on an experimental setup that comprises two standard cross-validations and a C++/Matlab toolkit to analyze surgical gestures employing hidden Markov models and utilizing linear dynamical systems. A summary of skill levels for different procedures of JIGSAWS is summarized in Table~\ref{table0}. As observed from the table the videos do not have the same number of frames and the number of videos for each task is limited.

\begin{table}
 \caption{ The summary of the JIGSAWS dataset}\label{table0}
  \centering
  \begin{tabular}{llll}
    \toprule
    \cmidrule(r){1-4}
    Label     & Sututring     & Knot Tying & Needle Passing \\
    \midrule
    Novel & 38 & 32 & 22\\
    Intermedite & 20 & 20  & 16 \\
    Expert & 20  & 20 & 18 \\ 
    \midrule
    Total Videos & 78 & 72 & 56\\
    Video Frames & 1794-9026 & 1014-2727 & 1789-4776\\
    \bottomrule
  \end{tabular}
  \label{tab:table}
\end{table}

\subsection{Gaussian Processes Classification}
Gaussian process (GP) is a non-parametric Bayesian approach for active learning and optimizing unknown functions for both regression and classification. Contrary to NN methods, GP is less dependent on the massive training data sets for generalization. This will help the development of an ASSA pipeline with limited training data. Since GP contains few training parameters, it is also computationally efficient. Thus, it does not add computational complexity when being used along with NN. In what follows, we will discuss the preliminaries for GP classification.

Suppose having a dataset as: \{$(\boldsymbol{x}_1,y_1), \cdots,(\boldsymbol{x}_N,y_N)$\} where $N$ is the number of data with $C$ different labels, such that: $y_i \in \{1,..., C\}$. 
For classification purpose it is desired to find latent function $F \in \mathcal{R}^{C \times N}$ such that $p(y_i |((f^1(\boldsymbol{x}_i),...,f^C(\boldsymbol{x}_i))))$ estimates the corresponding label of $\boldsymbol{x}_i$. For multi-class GPC where $C \geq 3$, $F$ can be written in the following form:

\begin{equation}
F=
\left (
\begin{array}{ccc}
\begin{array}{l}
f^1(\boldsymbol{x}_1)\\
f^2(\boldsymbol{x}_1)
\end{array}
& \cdots & 
\begin{array}{l}
f^1(\boldsymbol{x}_N)\\
f^2(\boldsymbol{x}_N)
\end{array} \\
\vdots & \ddots & \vdots\\
\begin{array}{l}
f^{c-1}(\boldsymbol{x}_1)\\
f^{c}(\boldsymbol{x}_1) 
\end{array} &
\cdots & 
\begin{array}{l}
f^{c-1}(\boldsymbol{x}_N)\\
f^{c}(\boldsymbol{x}_N)  \\
\end{array} 
\end{array}
\right )
\end{equation}

In multi-class GPC, contrary to Gaussian process regression and binary classification, more than one multivariate Gaussian distribution should be considered, such that for each label a multivariate Gaussian distribution should be defined as $\mathbf{f}_k \sim GP(0, K_k)$, while $\mathbf{f}_k = \{f^k(\boldsymbol{x}_1),...,f^k(\boldsymbol{x}_N)\}$, and $K_k$ is the relevant kernel function. Generally for estimating labels, \ref{eq2} can be used:

\begin{equation}
p(\mathbf{y}|F) = \prod\limits_{i=1}^{N}p\left(y_i |\left(f^1\left(\boldsymbol{x}_i\right),...,f^C\left(\boldsymbol{x}_i\right)\right)\right)
\label{eq2}
\end{equation}

Such a classification is computationally intense. Kim and Ghahremani~\cite{kim2006bayesian} introduced a method to reduce the complexity to $O(N^3)$. This complexity can be reduced to $O(MN^2)$ by Sparse Gaussian processes using pseudo-inputs~\cite{snelson2005sparse}. In several previous works, the right side of \ref{eq2} is replaced with a soft-max likelihood \cite{kim2006bayesian,williams1998bayesian} for multi-classification purposes, however, this is not scalable with the number of data instances, due to computational cost. As a result, \cite{hernandez2011robust} has presented a scalable multi-class GPC using the robust-max likelihood. Authors in \cite{villacampa2017scalable} have rendered a scalable multi-class GPC using expectation propagation by the Heaviside likelihood. However, these methods are still slow and computationally complicated.
To tackle complications, other solutions based on probabilistic data augmentation are introduced. For example, in~\cite{linderman2015dependent}, authors introduced a sequential method named Multinomial Stick Breaking structure using Pólya-gamma augmentation for further simplification of the computational complexity. 

\begin{figure*}
\centering
\subcaptionbox[]{\label{fig1a}}{\includegraphics[width=10cm]{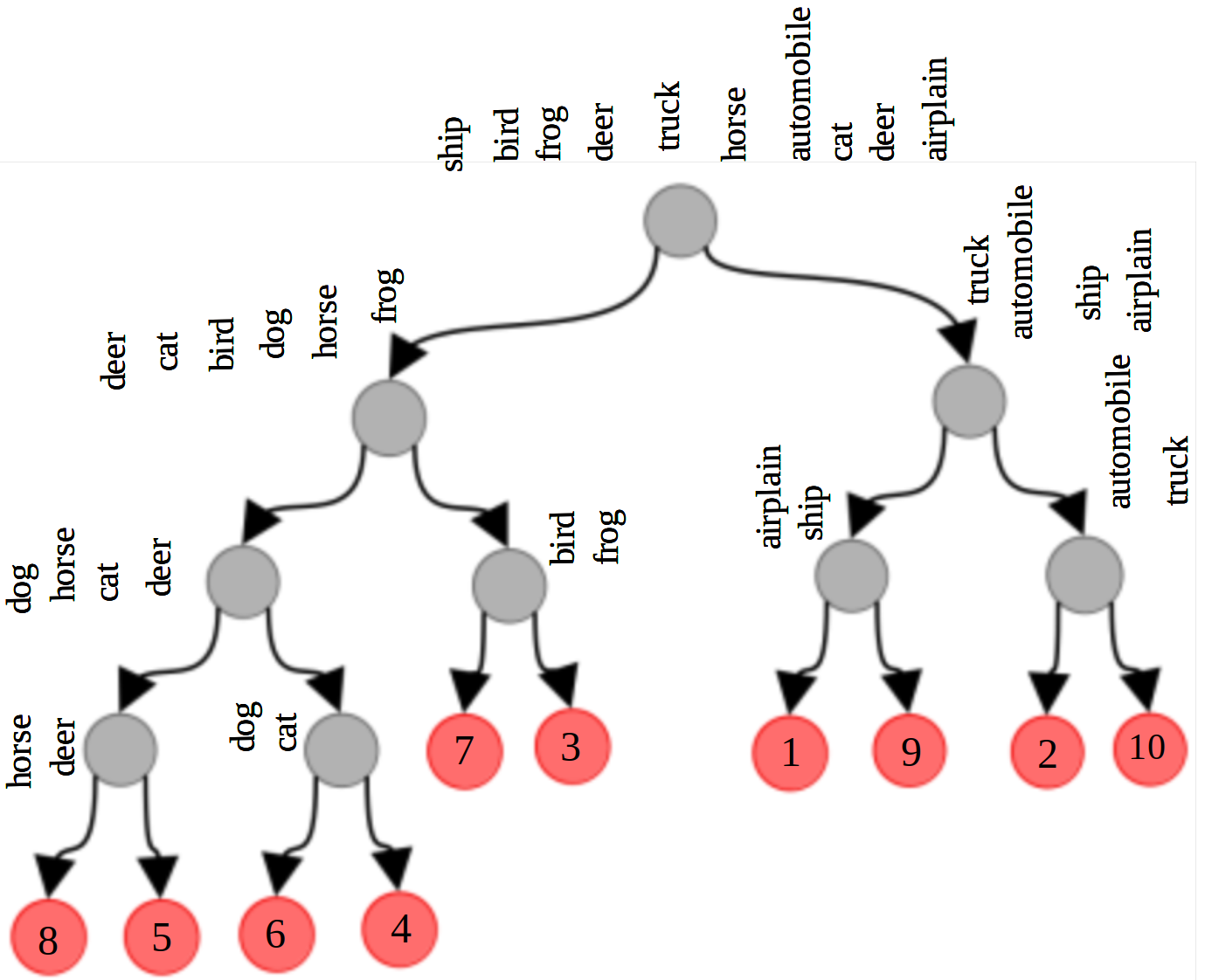}}%
\hfill
\subcaptionbox[]{\label{fig1b}}{\includegraphics[width=4cm]{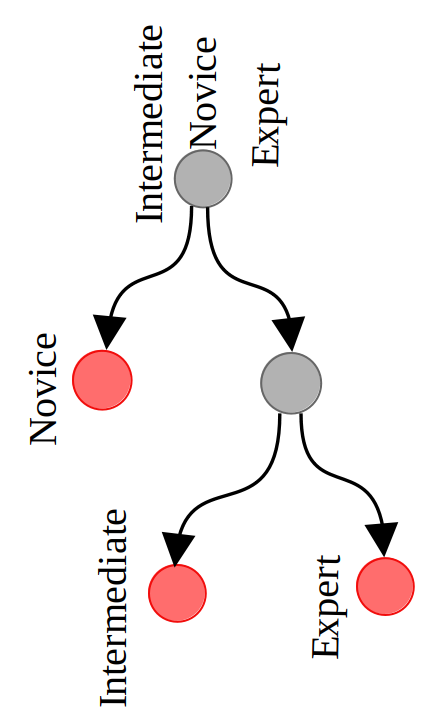}}%

\caption{(a) The structure of GP-Tree based on CIFAR-10~\cite{achituve2021gp},
(b) The structure of GP-Tree based on JIGSAWS.}
\end{figure*}

In another study, authors has introduced a hierarchy-based method to reduce the complexity of multi-class GPC, named GP-Tree for image classification using deep neural networks~\cite{achituve2021gp}. It is a tree-based model in which individual nodes leverage a sparse model of Gaussian processes by the Polya-Gamma augmentation to perform binary classification tasks. The structure of the tree is designed based on the number of labels. As an example, Figure~\ref{fig1a} illustrates the tree model for the CIFAR-10 data set. 
As observed from this figure, GP-Tree generates a balanced tree that is divided by the semantic meaning of the classes. For example for CIFAR-10 data, motorized vehicles are on the right subtree of the root node while animals are on the left one. This semantic partition is pronounced at all tree levels. One important aspect of this method is the architecture of the tree, where the authors discuss a method based on $kmeans++$ to assign labels to the nodes, which was previously introduced in~\cite{arthur2006k}. In this method, the latent functions of the GP, $\mathbf{f}_v$ for internal nodes $v$, can be obtained using:

\begin{equation}
\mathbf{f}_v \sim GP(\mathbf{m}_v , \mathbf{K}_v)
\end{equation}

in which $\mathbf{m}_v$ and $\mathbf{K}_v$ are the mean function and covariance function, respectively. 
Each GP can learn whether any input should go to the left child node or the right one in an unsupervised way. Two label arrays are allocated for each internal node (right $= 0$ and left $= 1$). Leaves are expressive of major labels, and for all the inputs from any class $c$, there is only a unique path from the root to the leaf of that class label ($P_c$), which results in estimating a label as follows:
\begin{equation}
p(y=c|\mathbf{F}) = \prod_{v \in P_c} \sigma(f_v)^{y_v}(1-\sigma(f_v))^{1-y_v}
\end{equation}

where $\mathbf{F}$ is a set of all the latent functions of the GP models in the tree and $\sigma$ is the logistic likelihood function. The GP model for each node uses Pólya-Gamma Data Augmentation~\cite{wenzel2019efficient} in which $X=(\boldsymbol{x}_1,...,\boldsymbol{x}_N) \in \mathbb{R}^{K \times N}$ are inputs of a GP and $\mathbf {y} = (y_1,...,y_N) \in \{0, 1\}^N$ are the labels which are outputs. Thus, the model can be defined as:

\begin{equation}
p(\mathbf{y,w,f,u},X) = p(\mathbf{y}|\mathbf{f,w})p(\mathbf{w})p(\mathbf{f}|\mathbf{u},X)p(\mathbf{u})
\label{eqq}
\end{equation}

In \ref{eqq}, $\mathbf w$ contains Pólya-Gamma variables, $\mathbf f$ is a vector of latent decision functions of $X$, and $\mathbf u$ indicates a vector of latent decision functions of inducing points.  Using the above equation, the prediction of the labels for the test data can be represented as:
\begin{equation}
  \begin{split}
   p(f_*|\boldsymbol{x}_*, \bar{\mathbf{X}}, \bar{\mathbf{y}}) \approx \int p(f_* | \mathbf{u},\boldsymbol{x}_*)q(\mathbf{u}) d\mathbf{f}\\
   = N(f_* | \mu_*,\scriptstyle \sum_*),\\
   \mu_* = \mathbf{k} _{m^*}^T \mathbf{K}^{-1}_{mm} \tilde{\mathbf{\mu}},\\
   \scriptstyle \sum_* = k_{**} - \mathbf{k}_{m^*}^T(\tilde{\mathbf{\scriptstyle \sum}}     \mathbf{K}^{-1}_{mm} - \mathbf{I})\mathbf{k}_{m^*} 
  \end{split}
\label{prediction}
\end{equation}

where $\bar{\mathbf{X}}$ denote the inducing points and $\bar{\mathbf{y}}$ are the labels~\cite{wenzel2019efficient}, $m$ is the number of inducing points.
In order to do inference in each node, variational inference has been employed. $q$ is the variational distribution. Additionally, $\mathbf{ K}_{mm}$ is the kernel for $m$ inducing points, while $*$ is used to represent test data instances. Notably, the computational complexity of any binary decision GP is $O(m^3)$. As a result, the computational cost of GP-Tree is about $O((C-1)m^3)$, which is substantially less than the general GPC.

\section{Proposed Method}
\label{proposed method}
\begin{figure*} [!ht]
\centering
\includegraphics[width=\linewidth]{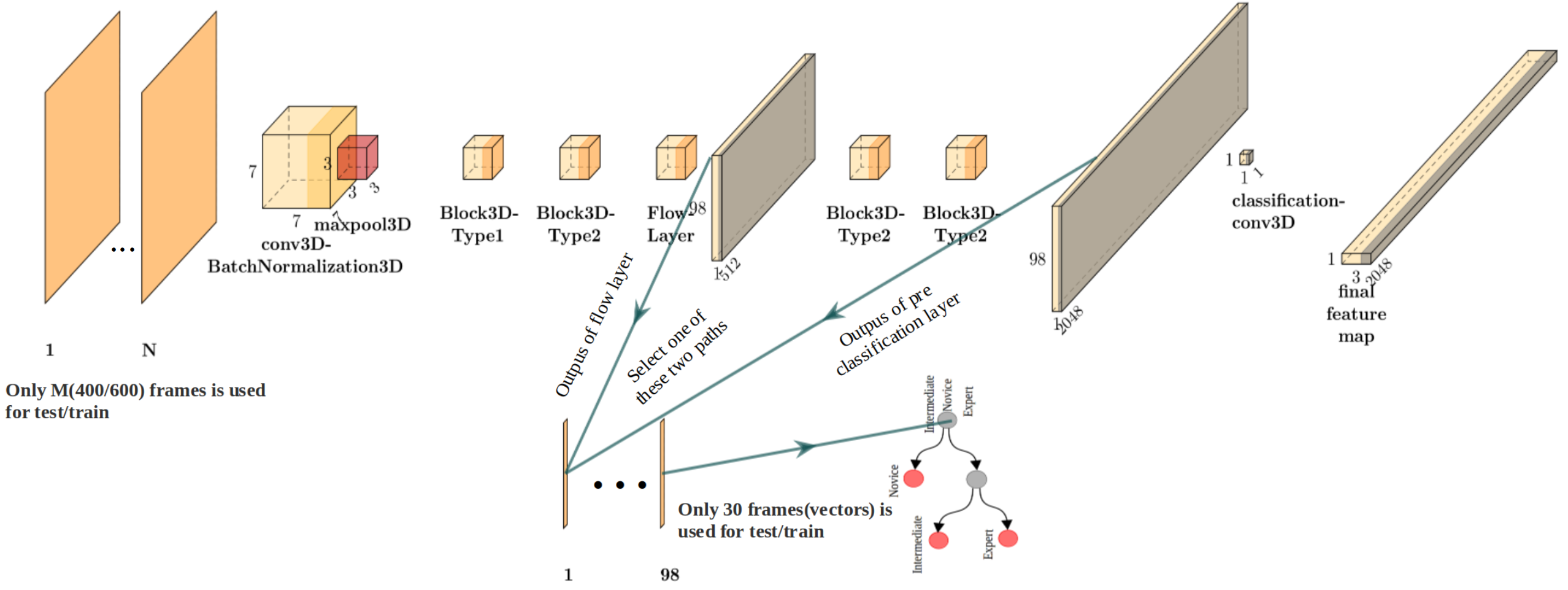}
\caption{The structure of the RF-based surgical skill assessment strategies. A drawn instance for a knot-tying video. The structure shows different combinations of the RF combined with different classification methods}
  \label{TS}
\end{figure*}

In this section, we elaborate on the proposed pipeline for ASSA using video data. The proposed method is a CNN with an inherent representation flow algorithm, originally introduced in~\cite{piergiovanni2019representation} coupled with NIGP-Tree (noisy input GP-Tree), namely RF-NIGP-Tree. As discussed earlier, NIGP-Tree is an enhanced GP-Tree method robust to noise, introduced in this study. The result of the analysis which will be discussed later in this paper,  shows that {RF-}NIGP-Tree improves the accuracy of ASSA. Furthermore, two variants of RBF kernel for GPC have been introduced. Compared to the state-of-the-art existing method~\cite{funke2019video}, the proposed method relies on fewer numbers of frames in training. Additionally, the number of frames for the test data has fewer frames compared to that of the train, while the utilized videos are resized. A schematic of the proposed pipeline is depicted in Figure~\ref{TS}.

In the proposed framework on contrary to the GP-Tree method introduced in~\cite{achituve2021gp}, the training of the CNN and NIGP-Tree classifier are independent. Since, The inputs of the NIGP-Tree classifier are subsequences of the feature maps, which are the outputs of the representation flow (RF) network. The feature map data are classified in both training and testing to be used for the training of the NIGP-Tree. The NIGP-Tree inherits the advantage of the GP method, which allows the train with limited data while considerably reducing computation. 
As discussed above, the inputs of the NIGP-Tree classifier are the feature maps of the CNN. Our analysis shows that these feature maps are noisy, which led us to introduce the NIGP-Tree classifier. In what follows, we elaborate on the development of the NIGP-Tree method along with newly introduced Kernels.

\subsection{NIGP-Tree}

As discussed above, feature maps are prone to noise, which could be because of the noisy inputs or the nature of the network. Motivated by this observation, we explore an enhanced GP-Tree method that is robust to noisy feature maps.

In~\cite{mchutchon2011gaussian, johnson2019accounting} authors have introduced robust-to-noise methods for Gaussian Process Regression and have mathematically proved the effectiveness of the proposed method and the robustness of the method when faced with noise. In~\cite{mchutchon2011gaussian} the GP model can be obtained using noisy data input, while in ~\cite{johnson2019accounting} data are firstly clear (without any noise) and are used for obtaining the GP model.
Based on these studies, we propose a robust-to-noise Sparse Gaussian process classification method using P{\`o}lya-Gamma data augmentation. Utilizing the same formulation as above studies and considering $X=[ \boldsymbol{x}_1,..., \boldsymbol{x}_n]\in \mathbb{R}^{K \times N}$ as the matrix of the denoised feature maps  of $S$:$\mathbf s_i = \boldsymbol{x}_i + \epsilon_{s_i}$, $\scriptstyle \sum_*$ in \eqref{prediction} can be written as:

\begin{center}
\[ \scalebox{1.25}{$\scriptstyle \sum_* = T_{**} + k_{**} - \mathbf k_{m^*}^T(\tilde{\mathbf \scriptstyle \sum}(\mathbf K^{-1}_{mm}) - \mathbf I)\mathbf k_{m^*}$} \]
\[ \scalebox{1}{$T_{ij} = T(\boldsymbol{x}_i, \boldsymbol{x}_j)=\partial _{u}(\boldsymbol{x}_i)(\scriptstyle \mathbf \sum_x)\textstyle \partial _{u}(\boldsymbol{x}_j)$} \]
\[ \scalebox{1}{$ \partial _{u}(\boldsymbol{x}_i) = \dfrac{\partial \mu_*}{\partial {\boldsymbol{x}_i}} $} \]
\end{center}

\setlength{\unitlength}{0.20mm}
where, $\mathbf \sum_x$ is the noise variance, which can be obtained by denoising the inputs. Having said this, we have employed a fast unsupervised method utilized in~\cite{pereyra2017fast} to accomplish this task. According to the mentioned works, this model can be trained and predicted using noisy data $S$ instead of $X$. Algorithm~\ref{alg1} summarizes the proposed NIGP-Tree method. 

\subsection{New Discovered Kernels}
Our study shows that for the JIGSAWS dataset, the RBF kernel results in a better performance compared to that of the other commonly used kernels. However, we have shown that the accuracy of the network can be significantly improved by modifying RBF kernel. Two modified RBF kernels are introduced in this regard. Considering the RBF kernel as follows:

\begin{equation}
k_{RBF}(\mathbf{\boldsymbol{x}_1,\boldsymbol{x}_2}) = \exp\Biggl(-1/2\left(\mathbf{\boldsymbol{x}_1}-\mathbf{\boldsymbol{x}_2}\right)^T\Theta^{-2}\left(\mathbf{\boldsymbol{x}_1}-\mathbf{\boldsymbol{x}_2}\right)\Biggr)
\label{rbf}
\end{equation}

where $\Theta$ is the length scale. Considering Eq.~\eqref{rbf} we introduce the following two kernels:

\begin{equation}
k_{30RBFs} = k_1 + k_2 + \cdots + k_{30}
\end{equation} 
\begin{equation}
k_{Powered-RBF}(\mathbf{\boldsymbol{x}_1,\boldsymbol{x}_2}) = \Bigl(k_{RBF} \Bigr)^\alpha
\end{equation}

It is to be noted that the latter one is computationally more efficient, in which $0.9<\alpha<2$ is a tuning hyperparameter. The performance of the classifier strategy using the proposed kernel is provided in detail in Section~\ref{results}.

\begin{algorithm}[!th]
\caption{NIGP-Tree}
\label{alg1}
\begin{algorithmic}[1]

\State $\mathbf S \leftarrow$ Noisy Feature Maps
\State $\mathbf L \leftarrow$ Labels
\State $\mathbf X \leftarrow$ Unsupervised-Feature-Map-Denoiser(S)
\State $\mathbf {Noise} \leftarrow (\mathbf S - \mathbf X)$
\State $\mathbf \sum_x \leftarrow variance(\mathbf {Noise})$
\State $\mathbf X \leftarrow$ select $\mathbf X$ or $\mathbf S$
\State select a kernel type$\{$RBF or 30RBFs or PowRBF$\}$
\State PS $\leftarrow$ the new presented prediction strategy
\State Build a NIGP-Tree
\State Set Hyperparameters
\State Initialize inducing locations $\bar{\mathbf X}$ and variational parameters
\For{epoch$ = 1,...,N$} 
    \State GO through any unique path related to each ($X$ and $L$)
    \For {any node}
        \State Update variational parameters 
    \EndFor
    \State Evaluate any node utilizing PS($\mathbf \sum_x$, $\mathbf X$ and $\bar{\mathbf  X}$)
    \State Update $\mathbf X$ and $\bar{\mathbf X}$ by the loss function of the total tree~\cite{achituve2021gp} using the computed variational lower bound for any node
\EndFor
\Return the Trained NIGP-Tree, $\mathbf X$, and $\bar {\mathbf X}$
\end{algorithmic}
\end{algorithm}

\section{Performance Analysis}
\label{Results}
In this section, the performance of the presented pipeline is investigated
for ASSA using the JIGSAWS dataset. Note that the proposed method is purely video-based and the assessment is done for three different tasks, suturing, knot tying, and needle passing. The levels of skills are differentiated with three labels, corresponding to novices, intermediate, and expert. For each task evaluation, the architecture of the network is the same while being trained separately. For both training and testing the frame size of the videos is scaled to be smaller. As discussed earlier, we compare the results to that of three different network architectures. In one architecture we employ a RF network. In the other two architectures, the RF is partially used combined with NIGP-Tree, while the input of the NIGP-Tree comes from different layers of the RF network. To show the effectiveness of the {RF-}NIGP-Tree, we have also had it compared with RF-GP-Tree. 
\subsection{Flow Architecture Representation}
The hyperparameters of the RF flow are that of the original network introduced in~\cite{piergiovanni2019representation} considering the non-fixed learning rate, initially set as $10^{-5}$. The batch size, momentum, weight-decay, and max-grad are also considered to be 1, 0.9, $10^{-6}$ and 100, respectively. In addition, for all the training, $\lambda$, $\theta$, and $\tau$ are the parameters of the flow layer in RF that have been trained, while keeping $w_x$ and $w_y$ fixed. It is based on the results of~\cite{piergiovanni2019representation} showing that fixed $w_x$ and $w_y$ will result in a better performance. For all the training the batch size is equal to 1, however, the numbers of epochs are considered to be 56, 95, and 112 for needle passing, suturing, and knot tying, respectively.
\subsection{Employing GP-Tree and NIGP-Tree}
As discussed earlier the input of the NIGP-Tree/GP-Tree classifier can either be the output of the flow layer or the output of the pre-classification layer of the RF, as depicted in Figure~\ref{TS}. The training of both NIGP-Tree and GP-Tree is unsupervised using the ELBO loss function, while for the test and validation, the cross-entropy loss function is utilized.
\subsection{Data preparation}
The original sizes of the videos for all three tasks in JIGSAWS are $640\times480$, which are scaled to $88\times88$ for suturing and $112\times112$ for both knot tying and needle passing. Furthermore, as it was represented in Table~\ref{table0}, JIGSAWS videos do not have equal lengths, and we have used a limited number of frames. The number of frames used for training are 1700, 600, and 900, and 900, 400, and 600 for the test, for suturing, knot tying, and needle passing, respectively. 

\subsection{Training and test process}
The output of the pre-classification layer are feature maps with $223 \times 2048$, $148 \times 2048$ and  $98 \times 2048$ sizes for suturing and needle passing and knot tying videos respectively. For reducing the computation time we have used the first 30 frames of each video for training and the second 30 frames are used for the test. As discussed for another study we have used the output of the flow layer with sizes of $223 \times 512 \times 11 \times 11$, $148 \times 512 \times 14 \times 14$, and $98 \times 512 \times 14 \times 14$. To reduce the complexity we just use the means of the last two dimensions, to be used with GP-Tree/NIGP-Tree. In this case, only the first 20 frames of each feature map are used for training and the second 20 frames are used for the test. 

\subsection{Results}
\label{results}
In this section, we summarize the performance of the proposed method for the three different models trained for each surgical skill assessment. We have carried out 12 different simulation studies as listed in Table.~\ref{tab2}, which also show an alternative name to simplify referencing the method. All the methods have RF as the core network with M1 being just the RF network. In M2-M7, RF is coupled with the GP-Tree classifier and in M8-M13 it is coupled with NIGP-Tree. Apart from M1 method the M2-M13 can be divided into six different categories based on the classifier that can be either GP-Tree or NIGP-Tree and the kernel that can be RBF, 30RBFs or PowRBF with different values for $\alpha$. Having said this for each category we provide the result of the best candidate. 

\begin{table}[!th]
\centering
\caption{List of methods and their alternative names} \label{tab2}%
\begin{tabular}{@{}lll@{}}
\toprule
Method Name & Alternative Name & category \\
\midrule
RF & M1 &N/A\\
RF-GP-Tree(RBF)   & M2 &1 \\
RF-GP-Tree(30RBFs) & M3 &2\\
RF-GP-Tree(PowRBF-$\alpha=0.9$) & M4 &3\\
RF-GP-Tree(PowRBF-$\alpha=1.3$) & M5 &3\\
RF-GP-Tree(PowRBF-$\alpha=1.5$) & M6 &3\\
RF-GP-Tree(PowRBF-$\alpha=1.9$) & M7 &3\\
RF-NIGP-Tree(RBF) &  M8 &4\\
RF-NIGP-Tree(30RBFs) & M9 &5 \\
RF-NIGP-Tree(PowRBF-$\alpha=.9$) & M10 &6\\
RF-NIGP-Tree(PowRBF-$\alpha=1.3$) & M11 &6\\
RF-NIGP-Tree(PowRBF-$\alpha=1.5$) & M12 &6\\
RF-NIGP-Tree(PowRBF-$\alpha=1.9$) & M13 &6\\
\bottomrule
\end{tabular}
\end{table}

In Table~\ref{table3} we have summarized the performance of the M1 method along with best candidate of each category for the classification of the suturing task. Studies are performed for classification using flow layer output and pre-classification layer output of the RF network. As observed from the accuracy and precision provided for each method, using a flow layer offers improved accuracy and precision. Comparing the results related to using flow layer output, shows that M3, M7, M9, and M13 have the same precision and accuracy for suturing task classification while M7 has the best computation time compared to the best candidate of all categories. It is notable that the training time of the RF provided in this table is not affecting the computation time as it is a pre-trained network and we have utilized the parameters obtained in \cite{piergiovanni2019representation}. As seen in Table~\ref{table4}, we have compiled the performance of the M1 technique and the top performers for each category in the classification of the needle-passing task. It is evident from the accuracy and precision values provided for each approach that using the flow layer output yields higher accuracy and precision. When comparing the results obtained from the flow layer output, we observe that M9 and M13 exhibit comparable precision and accuracy for needle passing task classification, while M13 displays a better computation time.

\begin{table}[!th]
\centering
 \caption{\label{tab:table-name}Classification results of suturing task using different methods. Precision = Recall = F1 Score}\label{table3}
\begin{tabular*}{\textwidth}{@{\extracolsep\fill}llclclclclclclclc}

\toprule%

\midrule
Method  & tr(s) & ts(s) & IPs & OS & LR & Pr & Acc\\
\midrule
M1   & 43200  &  437.19 & - & - & $10^{-5}$ &  0.83 &0.89 \\
\midrule
\multicolumn{8}{@{}c@{}}{Using pre-classification layer output} & \\
\midrule

M2 & 4.38 & 0.01 &  2 & 320 & 0.1 & 0.89 & 0.92 \\
M3 &  16.80 & 0.11  & 2 & 140 & 0.14 & 0.91 & 0.94\\
M6 & 4.59 & 0.009 & 2  & 220 & 0.2 & 0.92 & 0.946\\
M8 & 6.30 & 0.33 &  2 & 320 & 0.1 & 0.91 & 0.94\\
M9 & 17.18  & 0.41  & 2 & 175 & 0.18 & 0.92 & 0.95 \\
M12 & 6.21 & 0.32 & 2  & 260 & 0.16 & 0.94 & 0.96\\
\midrule
\multicolumn{8}{@{}c@{}}{Using flow layer output} & \\
\midrule
M2 & 3.80  & 0.13 & 2 & 200 & 0.2 & 0.92 & 0.95\\
M3 &  9.57 & 0.17  & 2 & 180 & 0.4 & 1.00 & 1.00\\
M7 & 3.52 & 0.12 &  2 & 80 & 0.6 & 1.00 & 1.00\\
M8 & 4.87 & 0.21 & 2 & 200 & 0.2 & 0.92 & 0.95\\
M9 & 10.78  &  0.24 & 2 & 180 & 0.4 & 1.00 & 1.00\\
M13 & 5.66 & 0.22 &  2 & 80 & 0.6 & 1.00 & 1.00\\
\bottomrule
\end{tabular*}
\footnotetext{tr(s): train time(seconds), ts(s):test time(seconds), IPs: inducing points, OS: output scale, LR: learning rate, Pr: precision, Acc: accuracy}

\end{table}

\begin{table}[!th]
\centering
 \caption{\label{tab:table-name}Classification results of Needle Passing task using different methods. Precision = Recall = F1 Score}\label{table4}
\begin{tabular*}{\textwidth}{@{\extracolsep\fill}llclclclclclclclc}

\toprule%

\midrule
Method  & tr(s) & ts(s) & IPs & OS & LR & Pr & Acc\\
\midrule
M1   & 18000  & 301.24  &  - &  - &$10^{-5}$ & 0.44 & 0.63\\
\midrule
\multicolumn{8}{@{}c@{}}{Using pre-classification layer output} & \\
\midrule

M2 &  4.23 &  0.53 &  2 &  20 & 0.25 & 0.87&0.91\\
M3 & 16.23  &  0.57 &  2 &  45 & 0.18 & 0.91 &0.94\\
M4 & 4.44  & 0.52 &  4 &  40 & 0.25 & 0.91 & 0.94\\
M8 & 7.30  &  0.54 &  2 &  40 & 0.25 & 0.91 & 0.94\\
M9 & 23.56  &  0.54 &  2 &  90 & 0.15 & 0.94 & 0.96\\
M10 & 8.45  & 0.52 &  3 & 250  & 0.1 & 0.92 & 0.95\\
\midrule
\multicolumn{8}{@{}c@{}}{Using flow layer output} & \\
\midrule
M2 & 3.70 & 0.12 & 2 & 180 & 0.18 & 0.91 & 0.94\\
M3 & 10.81  &  0.17 & 2 & 45 & 0.18 & 0.94 & 0.96\\
M7 & 3.82 & 0.13 &  2 & 45 & 0.18 & 0.94 & 0.96\\
M8 & 5.46  & 0.20  & 2 & 180 & 0.18 & 0.94 & 0.96\\
M9 & 12.72 & 0.24 & 2  & 45 & 0.18 & 0.982 & 0.988\\
M13 &  6.39 & 0.19  & 2 & 45 & 0.18 & 0.98 & 0.988\\
\bottomrule
\end{tabular*}
\footnotetext{tr(s): train time(seconds), ts(s):test time(seconds), IPs: inducing points, OS: output scale, LR: learning rate, Pr: precision, Acc: accuracy}

\end{table}

Table~\ref{table5} presents an overview of the performance of the M1 technique and the leading candidates in each category for the classification of the knot tying task. The provided accuracy and precision values demonstrate that utilizing the flow layer output leads to superior accuracy and precision. Upon comparing the results attained from the flow layer output, we conclude that M9 and M13 manifest comparable precision and accuracy for the knot tying task classification. Moreover, M13 showcases a superior computational time compared to M9.

\begin{table}[!th]
\centering
 \caption{\label{tab:table-name}Classification results of Knot Tying task using different methods. Precision = Recall = F1 Score}\label{table5}
\begin{tabular*}{\textwidth}{@{\extracolsep\fill}llclclclclclclclc}

\toprule%

\midrule
Method  & tr(s) & ts(s) & IPs & OS & LR & Pr & Acc\\
\midrule
M1   &  32400 & 543.49  & - & - & $10^{-5}$ & 0.67 & 0.79\\
\midrule
\multicolumn{8}{@{}c@{}}{Using pre-classification layer output} & \\
\midrule

M2 & 4.83 & 0.009 & 3 & 120 & 0.2 & 0.85 & 0.9\\
M3 & 18.13  &  0.04 & 4 & 60 & 0.25 & 0.87 & 0.91\\
M6 & 5.08 & 0.007 & 4 & 80 & 0.8 & 0.85 & 0.9\\
M8 &  6.37 & 0.28 & 4 & 250 & 0.25 & 0.92 & 0.95\\
M9 & 16.37 & 0.37 &  4 & 120 & 0.18 & 0.92 & 0.95\\
M11 & 6.51  &  0.29 & 6 & 220 & 0.23 & 0.92 & 0.95\\
\midrule
\multicolumn{8}{@{}c@{}}{Using flow layer output} & \\
\midrule
M2 & 3.59 & 0.14 & 3 & 250 & 0.2 & 0.87 & 0.91\\
M3 & 9.30  &  0.17 & 3 & 250 & 0.2 & 0.88 & 0.92\\
M7 & 3.62 & 0.13  &  2 & 250 & 0.2 & 0.87 & 0.91\\
M8 & 6.37  & 0.23  & 3 & 250 & 0.2 & 0.94 & 0.96\\
M9 & 13.61 & 0.2 & 3  & 250 & 0.2 & 0.98 & 0.99\\
M13 & 6.31  &  0.23 & 3 & 250 & 0.2 & 0.98 & 0.99\\
\bottomrule
\end{tabular*}
\footnotetext{tr(s): train time(seconds), ts(s): test time(seconds), IPs: inducing points, OS: output scale, LR: learning rate, Pr: precision, Acc: accuracy}

\end{table}

The results provided in Tables~\ref{table3} to \ref{table5} show the RF-NIGP-Tree provides better performance for classification of the surgical procedures. The comparison of the RF-NIGP-Tree with the state-of-the-art method in literature is provided in Table~\ref{table6}. A comparison of the results shows that our method has an average accuracy of 99\% while for the method of~\cite{funke2019video} and~\cite{tao2012sparse} the accuracy is 98.33 and 95.33 respectively. It is worth noting that the method of~\cite{funke2019video} utilizes a complex network that is computationally complex and also requires the full length of videos. On the other hand in~\cite{tao2012sparse}, both kinematic and video data are used, while our proposed method is purely video-based.

\begin{table}[!th]
\centering
 \caption{\label{tab:table-name}Some skill assessment approaches Results on JIGSAWS dataset}\label{table6}
\begin{tabular*}{\textwidth}{@{\extracolsep\fill}llclclclclclclclclc}

\toprule%
& \multicolumn{3}{@{}c@{}}{Suturing} & \multicolumn{3}{@{}c@{}}{Knot Tying} & \multicolumn{3}{@{}c@{}}{Needle Passing}\\

\midrule
Methods & Acc & F1.s & Rec & Acc & F1.s & Rec & Acc & F1.s & Rec\\
\midrule

3D Conv(RGB)   &  $1.00$ & $1.00$  & $1.00$ & $0.958 $ & $0.959 $ & $0.956$ & $0.964$ & $0.966$&$0.963$\\
3D-Conv(OF) & $1.00$ & $1.00$ & $1.00$ & $0.951 $ &  $0.950 $& $0.942 $ & $1.00$ & $1.00$&$1.00$\\
S-HMM & 0.97  & -  & - & 0.94 & - & - & 0.96& -&-\\
{RF-}NIGP-Tree & 1.00  & 1.00 & 1.00 & 0.99 & 0.98  & 0.98 &0.988 & 0.98 &0.98\\
\bottomrule
\end{tabular*}
\footnotetext{Acc: accuracy, F1.s: F1.score, Rec: recall, OF: optical flow}
\end{table}

\section{Conclusion}
\label{Conclusion}
In this article, a novel pipeline for surgical skill assessment is presented which can work with noisy videos and limited number of repetition of a certain task within the video. Using a same architecture we trained different models for evaluating each task and the results is compared with that of the literature. The comparison shows improvement on the average accuracy of surgical skill assessment for all the three tasks while being computationally simpler that literature methods.\\


\bibliographystyle{unsrt}  
\bibliography{references}

\end{document}